\documentclass[sigconf]{acmart}
\AtBeginDocument{%
  }

\acmConference[Under review]{ACM Conference}{Fifth ACM Symposium on Computer
Science and Law (2026)}{Berkeley, California}%
\acmISBN{978-1-4503-XXXX-X/2018/06}

\begin{document}

\title{Operationalising Extended Cognition: Formal Metrics for Corporate Knowledge and Legal Accountability}

\author{Elija Perrier}
\authornote{Under review for Fifth ACM Symposium on Computer
Science and Law (CS \& Law 2026).}
\email{elija.perrier@gmail.com}
\orcid{0000-0002-6052-6798}
\affiliation{%
  \institution{Centre for Quantum Software and Information, UTS}
  \city{Sydney}
  \state{NSW}
  \country{Australia}
}

\renewcommand{\shortauthors}{E. Perrier}

\begin{abstract}
 Corporate responsibility turns on notions of corporate \textit{mens rea}, traditionally imputed from human agents. Yet these assumptions are under challenge as generative AI increasingly mediates enterprise decision-making. Building on the theory of extended cognition, we argue that in response corporate knowledge may be redefined as a dynamic capability, measurable by the efficiency of its information-access procedures and the validated reliability of their outputs. We develop a formal model that captures epistemic states of corporations deploying sophisticated AI or information systems, introducing a continuous organisational knowledge metric (\(S_S(\varphi)\)) which integrates a pipeline's computational cost and its statistically validated error rate. We derive a thresholded knowledge predicate (\(\mathsf{K}_S\)) to impute knowledge and a firm-wide epistemic capacity index (\(\mathcal{K}_{S,t}\)) to measure overall capability. We then operationally map these quantitative metrics onto the legal standards of actual knowledge, constructive knowledge, wilful blindness, and recklessness. Our work provides a pathway towards creating measurable and justiciable audit artefacts, that render the corporate mind tractable and accountable in the algorithmic age.
\end{abstract}

\ccsdesc[300]{Computer systems organization~Distributed architectures}
\ccsdesc[500]{Information systems~Information retrieval}
\ccsdesc[500]{Information systems~Information systems applications}
\ccsdesc[500]{Social and professional topics~Management of computing and information systems}


\keywords{corporate knowledge, mens rea, extended cognition, algorithmic governance}



\maketitle

\section{Introduction}
The widespread adoption of powerful generative artificial intelligence (AI) systems for knowledge retention, retrieval and generation by corporations gives rise to important foundational questions regarding the attribution of knowledge to corporate entities. What a corporation knows - its state of knowledge - is central to the apportionment of rights and liabilities. Traditionally, a corporation’s culpability often depends upon its implied intentional state itself derived from the knowledge and collective intent of its human agents \cite{newyorkcentral1909,banknewengland1987} under doctrines such as the principle of \emph{respondeat superior}. These doctrines assume that corporate decision‑making - even when distributed across multiple agents - is intelligible to humans and that the epistemic horizon of the firm is limited to what its agents know. However the increasing adoption of large language models (LLMs) and large reasoning models (LRMs) - together with other forms of statistical analysis - problematises assumptions regarding knowing or understanding how those decisions are reached \cite{diamantis2020extended}. 

As companies integrate AI systems into their workflows, data processing and decision-making procedures, the scope of information that is readily and quickly available, via retrieval systems, prompt-based systems or agentic systems, grows considerably. For example, when a corporation deploys vector databases, implements retrieval-augmented generation pipelines, or integrates LLMs and/or LRMs across its infrastructure, it creates cognitive infrastructure analogous to human memory—but with radically different capacity constraints. This technological shift necessitates a re-evaluation of what knowledge itself means in a corporate context. We argue that corporate knowledge can no longer be seen as a static possession but must be understood as a dynamic capability, defined by two measurable dimensions: the \textit{efficiency} with which information can be accessed and the \textit{validated reliability} of the procedures that produce it. When the cost of discovering a fact drops from a month of archival research to milliseconds of computation, the legal standard for what a corporation knows, or ought to have known, must necessarily evolve.


Despite the important effects of AI upon corporate knowledge, there is currently no formal method for quantifying a corporation’s epistemic state when its cognition is distributed across humans and AI.  Existing legal approaches treat corporate knowledge as an objective quantity that can be ascertained \cite{asil2024machines}, yet they provide few if any metrics for measuring how AI expands a firm’s epistemic capacity or the role of algorithmic decision making per se \cite{bathaee2018blackbox,diamantis2020extended}. In an era where automated systems increasingly mediate or even make corporate decisions, understanding and measuring the bounds of what a corporation knows is paramount.
\subsection{Contributions}
In this paper, we address this methodological gap by proposing a formal, quantitative framework for corporate knowledge grounded in the principles of extended cognition and statistical learning. We set out to make corporate knowledge measurable in a way that is relevant to legal doctrine by applying metrics to the underlying search, generation, and verification procedures.

\begin{enumerate}
    \item We define corporate knowledge as a function of epistemic efficiency and procedural validity.
    \item We introduce an organisational knowledge metric, \(S_S(\varphi)\), that quantifies a firm's ability to establish a proposition \(\varphi\). This score integrates the expected computational cost (\(\mathbb{E}[T_\pi]\)) and the validated, end-to-end error rate (\(\varepsilon^{\mathrm{tot}}_\pi\)) of the optimal information-processing pipeline available to the firm. From this, we use computational science principles to derive a thresholded knowledge predicate, \(\mathsf{K}_S(\varphi;\theta_C)\), to formally impute knowledge. 
    \item We aggregate these findings into a firm-wide epistemic capacity index, \(\mathcal{K}_{S,t}\), a single metric that measures the weighted share of legally salient facts a corporation has the capacity to know at a given time. We show how specific technological improvements—such as deploying more efficient vector search or implementing more rigorous validation protocols—measurably increase this index, effectively and quantifiably expanding the corporate mind.
\end{enumerate}
The remainder of this paper is organised as follows. Section~\ref{sec:background} reviews the legal doctrines of corporate knowledge and the philosophical foundations of extended cognition, situating both within the context of contemporary AI technologies. Section~\ref{sec:formalmodel} develops a formal mathematical model for quantifying corporate knowledge. Section~\ref{sec:validation} then outlines the statistical validation methods that provide the evidentiary basis for this framework. Section~\ref{sec:mensrea} applies the model to legal standards of actual knowledge, constructive knowledge, wilful blindness, and recklessness. Finally, Section~\ref{sec:conclusion} discusses broader legal and governance implications and outlines directions for future research. Toy model simulation results illustrating the framework’s application are presented in the Appendix.

\section{Background \& Related Work}\label{sec:background}
The attribution of a mental state (\textit{mens rea}) to corporations is a foundational challenge in jurisprudence \cite{coffee1981nosoul}. Common law jurisdictions have traditionally solved this by imputing the knowledge and intent of human agents to the firm itself, primarily through underlying doctrines of agency, such as that of \textit{respondeat superior} \cite{newyorkcentral1909}. The need to impute an artifice of corporate epistemic states from its agents led to two main approaches distinguished by the nature, extent and criteria for aggregation of agent knowledge. The first is the identification doctrine where the corporate mind is attributed to its directing mind and will \cite{tesco1972,lennards1915,meridian1995}. The second is a more expansive collective knowledge doctrine, which in various forms has allowed for the aggregation of knowledge fragmented across multiple agents \cite{banknewengland1987}, where, in effect, the corporation is treated as a distributed information system. As with all legal concepts, these doctrines were conceived during an era when information technology was relatively rudimentary.  Although they have evolved in the decades since computational information technologies have risen to prominence, questions remain about the extent to which, in their current incarnations, they are well-suited to handle a modern corporation whose cognitive functions are deeply integrated with, augmented by and distributed across sophisticated algorithmic systems such as large language models, large reasoning models and agentic AI.

\subsection{Extended cognition}
One response to these questions can be found from the emerging disciplines of philosophy of mind. In particular, the theory of extended cognition, pioneered by Clark and Chalmers, provides both a philosophical and - as we demonstrate - practical lens for understanding this shift along with how to operationalise its consequences. The extended cognition thesis posits that cognitive processes can extend into the environment, incorporating external artefacts like notebooks or smartphones as functional parts of the mind \cite{clark1998extended,clark2025extending}. In effect cognition is framed as being scaffolded by memory and instrumental architecture that expands the cognitive capabilities, but also the site, of cognisant agents \cite{blitz2010freedom}. In this regard, it has much in parallel with emerging agentic AI systems founded upon LLMs that are situated within modern memory, retrieval and tool-use infrastructure \cite{chan2025infrastructure}. Legal scholars \cite{Bant_2023} have recently begun to apply the theory of extended cognition to corporations, arguing that a corporate infrastructure - such as servers, databases, and algorithms are not mere tools but constituent parts of its cognitive architecture \cite{diamantis2016corporate, diamantis2020extended}. As a result, they argue, a corporation's knowledge must therefore encompass what is functionally encoded within this infrastructure, rather than merely be ascertained by reference to some subset of its constituent agents \cite{cohen1935transcendental}. However, the nature of technical information infrastructure and computing has undergone a radical transformation since the extended cognition theory (and its precursors) were first developed. Current information systems integrating LLMs, generative AI and reasoning models have transcended the traditional reliable, static retrieval assumed by early extended mind theory. Today's corporate cognitive architecture is increasingly defined by a new class of information processing technologies fundamentally driven by underlying AI technologies. Examples include:
\begin{itemize}
    \item \textit{Dense Information Retrieval}. Corporations now organize vast, unstructured datasets (emails, reports, internal chats) into high-dimensional vector search spaces. Using Approximate Nearest Neighbor (ANN) search, they can retrieve semantically relevant information with a speed and efficiency—often in sub-linear time—that is orders of magnitude greater than traditional keyword search \cite{johnson2017faiss}. This dramatically increases the efficiency of epistemic access.
    \item \textit{Generative Knowledge Extraction}. Transformer-based Large Language Models (LLMs) do more than retrieve; they synthesize, summarize, and generate novel content from the data they are trained on. Technologies like Retrieval-Augmented Generation (RAG) combine dense retrieval with generation to produce contextualised answers, effectively allowing the corporation to in some sense interrogate its own data \cite{lewis2020rag}.
    \item \textit{Language Reasoning Models}. The frontier of this technology involves models capable of multi-step, "chain-of-thought" reasoning to solve complex problems \cite{lewis2021deal}. As these are deployed, the corporation's extended mind gains the capacity not just for memory and recall, but for cognitive labor.
\end{itemize}
These modern technological revolutions create a fundamental tension when it comes to assessing - and ascertaining - the nature and extent of corporate knowledge. While the efficiency of knowledge access has vastly increased, its reliability has become probabilistic. Unlike other deterministic systems, which (in a computational sense) faithfully retrieve information (e.g. an immutable notebook records what was written), sophisticated stochastic systems such as LLMs can inject greater uncertainty into the information retrieval process. LLMs are known to hallucinate, producing often apparently credible but factually incorrect information. The way in which LLMs and sophisticated models compare and assess truth is itself different from both deterministic models and methods typical within jurisprudential literature. Structured knowledge within knowledge graphs and the emergent logic of language reasoning models - such as chain-of-thought present new, complex forms of corporate cognition that defy simple legal analogies, it often being unclear whether for example reasoning traces contribute to model decisions, or reflect a conflation or superfluous artefact of model processing themselves. These differences between how information is retrieved and verified by comparison with traditional methods mean that the process of validation of generative outputs becomes even more important.  

\subsection{The need for a quantitative framework}
While as ever technology advances apace by comparison with the law, legal principles are sufficiently adaptable to address these developments. Analysis of existing legal doctrines shows that they already provide means of addressing the information consequences of emerging technologies. However, the nascency of generative AI and reasoning model technology means that the law often lacks the technical vocabulary to effectively evolve and adapt these principles of corporate knowledge. For example, the practical means by which a board's oversight duty to maintain reasonable information and reporting systems (as per  \textit{In re Caremark} \cite{caremark1996, stone2006ritter}) may require adaptation when interpreted in light of powerful informational \cite{fisse1983reconstructing} and reasoning models where almost unlimited knowledge and processing power is available instantaneously. Thus it may very well be the case that a board claiming ignorance of a fact that could be reliably discovered in milliseconds via a simple vector search is on far weaker ground than one that would have needed a team of archivists months to find the same information.
\\
\\
Similarly, epistemic doctrines by which liability is assigned \cite{clough2023failure}, such as wilful blindness, which requires a deliberate choice to avoid knowledge, may require updating. The fact that a firm avoided an inexpensive, high-speed query \cite{jewell1976, globaltech2011} which demonstrably would have obviated its ignorance may provide a new avenue for findings of wilful ignorance. The decision not to run such a query, or to avoid the standard validation procedures that would confirm the reliability of an AI system, becomes a quantifiable and deliberate epistemic choice. In each case, modern technology is profoundly reshaping the epistemic state, scope and extent of what may reasonably be regarded as corporate knowledge. This confluence of legal doctrine and technological change points to a central gap in jurisprudential literature: the absence of a formal method to measure and evaluate a corporation's epistemic state \cite{khanna1996corporate}. To say a corporation does know, or ought to have known, something requires a principled - and we argue measurable and falsifiable - way to assess its capacity to know \cite{list2011group}. 

\subsection{Operationalising the extended mind: efficiency and validation}
In this work, we propose such a framework. We argue that corporate knowledge in the AI era must be defined as a function of two measurable quantities: (i) the \textit{efficiency} of its information access procedures (retrieval and generation) and (ii) the \textit{validated reliability} of their outputs. 

\begin{enumerate}
    \item \textit{Epistemic efficiency (knowledge as speed of retrieval)}. The first limb of our conceptual approach is that knowledge is correlated to speed of retrieval or access. This first dimension concerns the cost of accessing knowledge, measured in time, computation, or other resources. For example, the difference between a linear search (\(O(n)\)) through terabytes of documents and a logarithmic-time vector search (\(O(\log n)\)) is not merely quantitative. It may reflect a qualitative transformation in what it means for information to be available or present to the corporate mind. Facts that were once practically undiscoverable, buried deep within the corporate data store, may now be rendered immediately accessible. This compression of access time means that the set of facts a corporation (and indeed its directors) might be reasonably expected to have present in mind has expanded dramatically. Our formal model, introduced in Section 3 below, sets out these concepts formally via the concept of information retrieval/generation pipelines, making epistemic efficiency a core component of the knowledge metric.
    \item \textit{Procedural validity: knowledge as reliable justified belief}.  The second dimension is the reliability of the process that produces a belief. The probabilistic and generative nature of modern AI means that the output of an information pipeline cannot be taken at face value. By this assumption, a belief is only justified - and can thus only count as corporate knowledge - if the procedure that generated it is demonstrably reliable. This opens the door for synthesising well-developed legal standards for expert evidence , such as those articulated in \textit{Daubert} and regulatory rules of evidence, directly with statistical and machine learning principles \cite{daubert1993, fre702_2023}. Thus the known or potential error rate required by courts in principle becomes a measurable quantity that can be estimated through rigorous statistical validation. As we will detail in Section 4, metrics such as cross-validated accuracy, calibration error, and generalisation bounds provide the evidentiary basis for certifying a knowledge pipeline's reliability. A corporation's failure to perform these validation steps, or its reliance on a pipeline with a known poor validation certificate, is a direct and measurable epistemic failure.
\end{enumerate}
As we set out below, by operationalising these concepts, we can arguably provide a technology-neutral and doctrinally grounded approach to quantifying the extended corporate mind in ways that are practically relevant for corporations and judicial practitioners. We begin with how measuring the extent and scope of a corporate epistemic state in terms of the efficiency by which knowledge may be retrieved or generated within a corporate information systems.


\section{Knowledge as Efficient Search: Formalising Corporate Epistemic States}\label{sec:formalmodel}

\subsection{From Epistemic Logic to Organizational Knowledge}
Classical epistemic logic treats knowledge as a modal operator on propositions, extended to groups via distributed and common knowledge—agents either know or do not know propositions \cite{fagin1995reasoning,halpern1990knowledge}. This binary framework proves inadequate for corporations whose cognition distributes across human agents and algorithmic systems with varying degrees of reliability. We instead adopt a continuous knowledge metric aligned with legal standards of proof. In corporate settings, doctrines like "collective knowledge" already ascribe a firm's epistemic state to the aggregation of information across agents \cite{banknewengland1987}. Building on these princples, we treat a corporation's knowledge as the outcome of procedures that efficiently retrieve and reliably verify information available across its people and technical substrates (databases, logs, retrieval indices, language models, and verification tools), consistent with doctrines such as wilful blindness and constructive knowledge \cite{jewell1976,giovannetti1990,globaltech2011,mpc1985}.

\subsection{A Search-Generation-Verification Model of Corporate Knowledge}
Let \(\mathcal{U}\) be a universe of propositions, \(\Phi \subseteq \mathcal{U}\) the legally relevant facts at time \(t\), and
\[
S \;=\; (D, I, \mathcal{R}, \mathcal{G}, \mathcal{V}, \Pi, \Theta)
\]
a corporate cognition stack comprising data stores \(D\), indices \(I\) (e.g., vector stores), retrieval procedures \(\mathcal{R}\), optional generators \(\mathcal{G}\) (e.g., for RAG synthesis), verifiers \(\mathcal{V}\), a set of admissible pipelines \(\Pi\) (which are compositions of retrieval/generation and verification), and policy parameters \(\Theta = (\tau^\star, ...)\) that set a reference time scale.

For each pipeline \(\pi\in\Pi\) seeking to establish a proposition \(\varphi\in\Phi\), let \(\mathbb{E}[T_\pi]\) denote its expected cost (e.g., time or query budget). We define three component error rates: retrieval error \(\varepsilon^{\mathrm{ret}}_\pi\) (miss/false-hit), generation error \(\varepsilon^{\mathrm{gen}}_\pi\) (hallucination/misrepresentation, set to 0 for pure retrieval pipelines), and verification error \(\varepsilon^{\mathrm{ver}}_\pi\) (Type I/II). We compose these using a multiplicative formula for total error:
\begin{equation}
\varepsilon^{\mathrm{tot}}_\pi \;=\; 1 - (1-\varepsilon^{\mathrm{ret}}_\pi)(1-\varepsilon^{\mathrm{gen}}_\pi)(1-\varepsilon^{\mathrm{ver}}_\pi).
\label{eq:total_error}
\end{equation}
where our composition assumes independence (non-correlated errors). Where component errors are dependent, an empirical joint error or an appropriate bound should be used.

\subsection{The Continuous knowledge metric and Predicate}
To capture both efficiency and reliability, we first model the cost of knowledge acquisition via a monotone decreasing function \(f_\tau(x) = (1 + x/\tau^\star)^{-1} \in (0,1]\), where \(\tau^\star\) is a context-specific reference time. This allows us to define a continuous score for each individual pipeline:
\begin{equation}
s_\pi(\varphi) \;=\; f_\tau\!\big(\mathbb{E}[T_\pi]\big)\cdot \big(1 - \varepsilon^{\mathrm{tot}}_\pi\big).
\label{eq:pipeline_score}
\end{equation}
The \textit{organisational knowledge metric} $S_S(\varphi)$ about a proposition \(\varphi\) is then the best achievable score across all available pipelines, representing the corporation's maximal epistemic access to \(\varphi\):
\begin{equation}
S_S(\varphi) \;=\; \sup_{\pi\in\Pi}\; s_\pi(\varphi)\; \in [0,1].
\label{eq:org_score}
\end{equation}
Finally, to connect this continuous measure to legal thresholds, we introduce a context-dependent parameter \(\theta_C\in(0,1)\) that reflects the stakes or relevant legal standard. This yields the thresholded knowledge predicate above which the corporation is deemed to have knowledge of \(\varphi\):
\begin{equation}
\mathsf{K}_S(\varphi;\theta_C) \;=\; \mathbf{1}\!\left\{\, S_S(\varphi)\;\ge\; \theta_C \right\}.
\label{eq:predicate_threshold}
\end{equation}
This framework has two key properties. First is \textit{composition}: if a retrieval step \(\pi_1\) has cost \(\mathbb{E}[T_{\pi_1}]\) and error \(\varepsilon^{\mathrm{ret}}_{\pi_1}\), and a verification step \(\pi_2\) has cost \(\mathbb{E}[T_{\pi_2}]\) and error \(\varepsilon^{\mathrm{ver}}_{\pi_2}\), their composition \(\pi=\pi_2\circ\pi_1\) has an additive cost and a composed error. Second is \textit{monotonicity}: if a new technology (e.g., a better index) uniformly reduces the cost or error of a pipeline, the organisational knowledge metric \(S_S(\varphi)\) increases or stays the same, weakly expanding the set of propositions for which the knowledge predicate holds.

\subsection{Firm-Level Epistemic Capacity and the Epistemic Frontier}
To assess a corporation's overall knowledge state across the set of \(\varphi\), we aggregate the scores of interest. Let \(w:\Phi\to\mathbb{R}_{\ge 0}\) be a weighting function that encodes the salience of each proposition to a subject matter the corporate knowledge of which we wish to measure. We can define an \textit{epistemic capacity index} \(\mathcal{K}_{S,t}\) as the weighted fraction of salient facts that the corporation can know, given its current capabilities and a set of thresholds \(\theta(\varphi)\):
\begin{equation}
\mathcal{K}_{S,t}(\theta) \;=\;
\frac{\sum_{\varphi\in\Phi} w(\varphi)\,\mathbf{1}\{\, S_S(\varphi)\ge \theta(\varphi)\,\}}
{\sum_{\varphi\in\Phi} w(\varphi)} \;\in\; [0,1).
\label{eq:capacity_index}
\end{equation}
We can characterise a corporation's knowledge in terms of an epistemic frontier, \(\mathscr{F}_S=\{\,(\mathbb{E}[T_\pi],\varepsilon^{\mathrm{tot}}_\pi): \pi\in\Pi\,\}\), which is the set of all achievable cost-error pairs. Technological or organisational improvements, such as the use of generative AI, LLMs, reasoning models or agents, can be seen - if effective - to shift this frontier outward, lowering costs and errors and thereby increasing the firm's epistemic capacity \cite{johnson2017faiss,guo2017calibration,lewis2020rag}.

\subsection{Instantiation with Modern Retrieval and AI}
This formal framework can be implemented with practical corporate AI pipelines. A common example is a system using Retrieval-Augmented Generation (RAG). RAG systems are a form of integrated vector database and search scaffolding (usually over LLMs). They involve multiple components: dense retrieval over a vector index \(I\) using approximate nearest neighbors (\(\mathcal{R}\)); evidence synthesis using a language model (\(\mathcal{G}\)); and verification via cross-source checks or a calibrated classifier (\(\mathcal{V}\)). Retrieval recall \(r\) corresponds directly to an error term, \(\varepsilon^{\mathrm{ret}} = 1-r\). If we assume no generation error (\(\varepsilon^{\mathrm{gen}}=0\)), the total error is bounded by \(\varepsilon^{\mathrm{tot}}_\pi \leq 1-r(1-\varepsilon^{\mathrm{ver}})\). The cost, \(\mathbb{E}[T_\pi]\), scales with the index probe budget [\textbf{query cost}] plus the computational time for synthesis and verification. The logs of queries, retrieved documents, prompts, confidence scores, and verification decisions become the essential audit artefacts that can support (or undermine) a corporate claim regarding its knowledge state \cite{kroll2017accountable,bathaee2018blackbox}. This formalisation has several consequences for legal doctrine and corporate governance. First, it provides a measurable means of identifying precisely how ingesting corporate data with AI expands the corporate mind: improving the components of the stack \(S\) (e.g., by adding a new index \(I\) or retriever \(\mathcal{R}\)) pushes the epistemic frontier \(\mathscr{F}_S\) outward, thereby increasing the organizational knowledge metric \(S_S(\varphi)\) and widening the set of propositions for which \(\mathsf{K}_S\) holds. Second, in principle, while not usually quantified, compliance frameworks and regulatory guidance could implicitly define reference thresholds \(\theta(\varphi)\). Prosecutors and regulators could then evaluate whether a firm's failure to establish a fact was due to a genuine lack of capacity or a failure to execute an available, high-scoring pipeline. Third, doctrines of extended and functional corporate knowledge fit naturally within this model. When algorithmic components perform epistemic work under firm control, their procedural capabilities and outputs properly contribute to the corporate knowledge state \cite{diamantis2020extended}. Finally, the framework provides an additional basis for assessing culpability such as wilful blindness. The deliberate suppression of a cheap or cost-effective test e.g. a pipeline \(\pi\) with a very high score \(s_\pi(\varphi)\) due to low cost and low error - could be regarded as providing evidence of corporate intentionality relevant to \textit{mens rea} \cite{jewell1976,giovannetti1990,globaltech2011}.

\section{Validation metrics and statistical learning}\label{sec:validation}
The formal framework introduced in Section 3 provides a structure for corporate knowledge, but its legal utility depends on grounding its parameters—particularly the error term \(\varepsilon^{\mathrm{tot}}_\pi\)—in evidence that is both statistically sound and judicially cognizable. This section operationalizes the abstract model by translating its components into concrete, verifiable metrics derived from statistical learning theory. We demonstrate how established validation techniques can produce a justiciable record of a pipeline’s reliability, creating an evidentiary basis for applying the knowledge predicate \(\mathsf{K}_S(\varphi;\theta_C)\) and calculating the epistemic capacity index \(\mathcal{K}_{S,t}\). The objective is to construct an evidentiary package for each pipeline \(\pi\) that quantifies its expected cost, its out-of-sample error rate, and the trustworthiness of its probabilistic outputs, thereby providing a principled answer to whether the corporation knows \(\varphi\).

\subsection{Reliability, Error Rates, and the Validation Certificates}
Sophisticated AI and information retrieval systems are analogous to expert systems - both in terms of domain expertise, but also the capacity for information processing. We can draw upon the jurisprudential standards for expert evidence in order to obtain practical metrics according to which validation of corporate knowledge may occur. The jurisprudence of scientific and technical evidence, articulated in the \textit{Daubert} sequence of cases and codified in Federal Rule of Evidence 702, requires that any expert methodology be testable, possess a known or potential error rate, and be reliably applied to the facts of the case \cite{daubert1993,kumho1999,fre702_2023}.  We can directly address these legal requirements using standard statistical guarantees. For any predictive model \(f\) within a pipeline \(\pi\) (whether used for retrieval, generation, or verification), we can select a bounded loss function \(\ell\in \{0,1\}\) (e.g., 0-1 loss for classification) and compute its empirical risk on held-out or cross-validated data: \(\widehat{L}_n(f)=\tfrac{1}{n}\sum_{i=1}^n \ell(f(x_i),y_i)\). One example of a standard concentration inequality, such as Hoeffding's, which provides a high-confidence upper bound on the true, unobservable risk \(L(f)\):

\begin{equation}
\Pr\!\Big(L(f)\le \widehat{L}_n(f)+\sqrt{\tfrac{\ln(1/\delta)}{2n}}\Big)\ \ge\ 1-\delta.
\label{eq:hoeff}
\end{equation}
where \(L(f)\) is the model's true error rate on the underlying data distribution, \(\widehat{L}_n(f)\) is the measured error rate on a test sample of size \(n\), and \(1-\delta\) is the desired statistical confidence level. This inequality allows us to translate an observed error rate into a conservative, high-probability estimate of worst-case future performance. To create a justiciable summary of a pipeline's performance, we introduce the concept of a \textit{validation certificate}, which collects these conservative, high-confidence bounds for each component:

\begin{equation}
\mathsf{Cert}_\delta(\pi)=\Big(\widehat{\mathbb{E}}[T_\pi],\; U_\delta(\varepsilon^{\mathrm{ret}}_\pi),\; U_\delta(\varepsilon^{\mathrm{gen}}_\pi),\; U_\delta(\varepsilon^{\mathrm{ver}}_\pi)\Big),
\label{eq:cert}
\end{equation}
where \(\widehat{\mathbb{E}}[T_\pi]\) is the measured average cost and each \(U_\delta(\cdot)\) is a conservative upper bound on the error of a component, derived from an appropriate statistical test (e.g., Eq. \ref{eq:hoeff} or a binomial confidence interval like the Wilson score interval). By substituting these upper bounds into our total error formula (Eq. \ref{eq:total_error}), we obtain a conservative upper bound on the total pipeline error, \(U_\delta(\varepsilon^{\mathrm{tot}}_\pi)\). This allows for a direct, high-confidence test of the knowledge predicate:
\begin{equation}
f_\tau\!\big(\widehat{\mathbb{E}}[T_\pi]\big)\cdot \big(1-U_\delta(\varepsilon^{\mathrm{tot}}_\pi)\big)\ \ge\ \theta_C.
\label{eq:plug_test}
\end{equation}

If this inequality holds, we can assert with confidence at least \(1-\delta\) that the pipeline \(\pi\) meets the epistemic threshold for knowing \(\varphi\). This transforms a statistical finding into a legally relevant conclusion: the procedure is demonstrably efficient and reliable enough to constitute corporate knowledge.

\subsection{Calibration}
For a model's output to be legally actionable, its claimed confidence must correspond to its actual reliability. Raw probability scores from a machine learning model are not, by themselves, beliefs that the law should credit. Instead, calibration of models is often undertaken in order to transform them into reliable inputs into decisional workflows. A verifier or generator is well-calibrated if its probabilistic predictions align with empirical frequencies. For example, when it asserts 80\% confidence, it is correct in approximately 80\% of such cases. Miscalibration can be quantified using metrics like the Expected Calibration Error (ECE), which measures the weighted average difference between a model's predicted confidence and its observed accuracy across bins of predictions. To compute ECE, predictions are grouped into \(M\) bins \( (B_1, ..., B_M) \) based on their confidence scores. The ECE is then the weighted average of the difference between the average accuracy (\(\mathrm{acc}(B_k)\)) and the average confidence (\(\mathrm{conf}(B_k)\)) within each bin:
\begin{equation}
\widehat{\mathrm{ECE}}=\sum_{k}\frac{|B_k|}{n}\,\big|\mathrm{acc}(B_k)-\mathrm{conf}(B_k)\big|.
\label{eq:ece}
\end{equation}
Poor calibration is often a remediable flaw. Post-hoc recalibration techniques (such as scaling) can frequently reduce the \(\widehat{\mathrm{ECE}}\) and improve a model's reliability without the need for complete retraining \cite{guo2017calibration,platt1999}. From a legal standpoint, a low ECE provides the confidence corporations and courts may require for relying on a model's outputs. It allows a corporation to set decision thresholds (e.g., "initiate review if predicted risk \(P\ge\gamma\)”) on a principled basis, as the resulting error rates are transparent and correspond to the model's stated confidence. Calibrated confidence is what transforms a model's numerical output into a belief that the corporation can reasonably justify.

\subsection{Generalisation}
A model’s reliability is only meaningful if it generalises from training data to unseen, real-world scenarios. The legal requirement that a method be reliably applied demands that the validation process mirror the anticipated deployment context. If data is independent and identically distributed, \(K\)-fold cross-validation is a standard and robust approach. However, if data has a temporal structure (e.g., financial transactions), a rolling-window or forward-chaining validation is necessary to prevent testing on future data. Similarly, if data is clustered (e.g., by user or location), group-aware validation is required to ensure the model generalizes across clusters \cite{kohavi1995study,bergmeir2018note}. The cross-validated risk:
\begin{equation}
\widehat{L}_{\mathrm{CV}}=\frac{1}{K}\sum_{j=1}^K \frac{1}{|V_j|}\sum_{i\in V_j}\ell\big(f_{-j}(x_i),y_i\big),
\label{eq:cv}
\end{equation}
where \(f_{-j}\) is the model trained without fold \(j\), provides an unbiased estimate of out-of-sample performance. This is the figure of merit that could in principle be relevant to a court or regulator. To prevent overfitting and avoid reliance on spurious correlations—a form of "reliably wrong" reasoning—model selection should be disciplined by penalizing complexity. We therefore select verifiers according to a rule that balances performance with parsimony. Here, the dataset is partitioned into \(K\) disjoint folds, \(V_1, ..., V_K\). For each fold \(V_j\), a model \(f_{-j}\) is trained on the remaining \(K-1\) folds and then evaluated on \(V_j\). The cross-validated risk is the average risk across all folds:
\begin{equation}
\widehat{f}\in\arg\min_{f}\ \widehat{L}_{\mathrm{CV}}(f)+\lambda\,\Omega(f),\qquad \Omega\in\{\mathrm{AIC},\mathrm{BIC},\mathrm{MDL}\}.
\label{eq:penal}
\end{equation}
Here, \(\Omega(f)\) is a penalty term that measures the complexity of the model \(f\) (e.g., using criteria like AIC, BIC, or MDL), and \(\lambda\) is a regularization parameter that controls the trade-off between the model's out-of-sample error and its complexity \cite{akaike1974,schwarz1978,rissanen2007,guo2017calibration,nist_rmf_2023,eu_ai_act_2024,sr11_7_2011}. By incorporating a complexity penalty, we ensure that additional model complexity is only accepted if it provides a commensurate improvement in out-of-sample performance \cite{akaike1974,schwarz1978,rissanen2007}. In essence, our validation protocol must reflect the realities of deployment, and our model selection must reward elegant simplicity over un-generalisable complexity.

\subsection{Integration with the Corporate Knowledge Framework}
With the validation certificate \(\mathsf{Cert}_\delta(\pi)\) established, the final step is to integrate these statistically-grounded metrics back into our legal-epistemic framework. To do so, we want to compute a lower-bound estimate of the pipeline score:
\begin{equation}
s^{\mathrm{LB}}_\pi(\varphi)=f_\tau\!\big(\widehat{\mathbb{E}}[T_\pi]\big)\cdot\big(1-U_\delta(\varepsilon^{\mathrm{tot}}_\pi)\big).
\label{eq:scorelb}
\end{equation}
By taking the supremum of this score over all available pipelines, \(S^{\mathrm{LB}}_S(\varphi)=\sup_{\pi\in\Pi}s^{\mathrm{LB}}_\pi(\varphi)\), we can declare knowledge if \(S^{\mathrm{LB}}_S(\varphi)\ge \theta_C\). This serves as a high-confidence surrogate for the knowledge predicate in Eq. \ref{eq:predicate_threshold}, providing a defensible assertion that conditional upon estimated resource constraint and error, the knowledge retrieval/generation system satisfies the threshold for knowledge. By inserting this high-confidence indicator into Eq. \ref{eq:capacity_index}, we obtain a lower-bound estimate of the epistemic capacity index, \(\underline{\mathcal{K}}_{S,t}\). This metric represents the weighted share of salient facts that the firm could have timely known, at a statistical confidence level of \(1-\delta\), using its validated and calibrated AI pipelines. In effect, we thereby operationalise the corporate epistemic state in a way that is potentially utilisable by corporations, regulators and courts when determining corporate knowledge and any consequential liability. Although simplified, this model presents potential direct and defensible pathway from statistical measures of information to legal imputation of corporate knowledge.

\section{Relating search knowledge to legal mens~rea}\label{sec:mensrea}
Our formal model of corporate knowledge provides a vocabulary for interpreting and applying traditional standards of \textit{mens rea}. Legal doctrines of intent often hinge on what a corporation knew or should have known. By connecting these doctrines to the organizational knowledge metric (\(S_S(\varphi)\)), the resulting knowledge predicate (\(\mathsf{K}_S(\varphi;\theta_C)\)), and the firm-wide epistemic capacity index (\(\mathcal{K}_{S,t}\)), we can create a more rigorous and technology-neutral framework for attributing mental states. The legal question of what a "reasonable corporation" would do can be framed as setting a context-dependent knowledge threshold, \(\theta_C\), which dictates when knowledge is formally imputed.

\subsection{Actual knowledge} Actual knowledge is established when a corporation not only possesses the capacity to know a fact but has actually executed a procedure to confirm it. In our framework, this means a pipeline \(\pi\) was run, and its validated performance was sufficient to meet a high standard of reliability. Formally, a claim of actual knowledge of a proposition \(\varphi\) is substantiated when the corporation has executed a pipeline \(\pi\) whose conservative, lower-bound score meets a high threshold for actual knowledge (\(\theta_{AK}\)). This successful validation allows the corporation to assert that, for this specific proposition, the knowledge predicate is met:
\[ s^{\mathrm{LB}}_\pi(\varphi) = f_\tau\!\big(\widehat{\mathbb{E}}[T_\pi]\big)\cdot\big(1-U_\delta(\varepsilon^{\mathrm{tot}}_\pi)\big) \ge \theta_{AK} \quad \implies \quad \mathsf{K}_S(\varphi;\theta_{AK}) = 1. \]
The two core components of our approach - knowledge as efficient, validated access - are both satisfied. The efficiency term, \(f_\tau(\cdot)\), confirms that the information was readily accessible to the corporate mind, akin to being present to mind. The reliability term, \((1-U_\delta(\varepsilon^{\mathrm{tot}}_\pi))\), confirms that the accessed information is not mere speculation but is a belief warranted by a statistically sound procedure. The execution of this pipeline produces a result that can be treated as a corporate admission or a business record, making the knowledge concrete and attributable. The auditable validation certificate, \(\mathsf{Cert}_\delta(\pi)\) (Eq. \ref{eq:cert}), provides the evidentiary basis for this assertion. 
This certificate is the critical link between our model and the requirements of U.S. evidence law. Under the \textit{Daubert} standard and Federal Rule of Evidence 702, expert testimony (including that derived from algorithmic systems) must be the product of reliable principles and methods that are reliably applied \cite{daubert1993, fre702_2023}. The validation certificate serves as a pre-packaged evidentiary record that directly addresses these requirements:
\begin{itemize}
    \item The known or potential error rate required by \textit{Daubert} is arguably satisfied by the high-confidence upper bounds on error, \(U_\delta(\varepsilon^{\cdot}_\pi)\), which are themselves derived from the rigorously computed cross-validated risk, \(\widehat{L}_{\mathrm{CV}}\) (Eq. \ref{eq:cv}).
    \item The reliability is satisfied by documenting that the validation strategy matched the deployment context (e.g., temporal splits for time-series data) and that the model's probabilistic outputs were trustworthy, as confirmed by a low ECE (Eq. \ref{eq:ece}).
    \item The justification for the chosen method is supported by documentation of model selection using complexity penalties like \(\Omega(f)\) (Eq. \ref{eq:penal}), demonstrating that the chosen pipeline was not unnecessarily complex or prone to overfitting.
\end{itemize}
This approach aligns with judicial standards requiring demonstrated reliability, as seen in cases like \textit{United States v. Bonallo}, and provides a formal basis for inferring knowledge from internal corporate records \cite{bonallo1988, banknewengland1987}. Consider two concrete examples.
\paragraph{Antitrust Compliance} An energy company uses an algorithmic trading system. To ensure compliance, it runs a daily pipeline (\(\pi_{comp}\)) designed to test the proposition (\(\varphi\)): "Our trading algorithm's bids in the last 24 hours were not statistically correlated with the bids of Competitor X." The pipeline retrieves market data, runs a pre-validated time-series analysis, and produces a report. The \(\mathsf{Cert}_\delta(\pi_{comp})\) documents a 10-minute runtime (\(\widehat{\mathbb{E}}[T_\pi]\)), a 99\% accuracy rate in identifying collusive patterns in historical simulations (\(U_\delta(\varepsilon^{\mathrm{ver}}_\pi) = 0.01\)), and robust calibration. When the pipeline runs and confirms no correlation, its high score, \(s^{\mathrm{LB}}_{\pi_{comp}}(\varphi) \ge \theta_{AK}\), means the corporation has actual knowledge of its compliance for that day. The outputted report is a definitive business record that can be used to defend against an accusation of price-fixing.

\paragraph{Product Liability Monitoring} A medical device manufacturer has a pipeline (\(\pi_{safe}\)) that continuously monitors real-world performance data from its devices to test the proposition (\(\varphi\)): "Device Model Y is operating within its specified safety tolerance of <0.05\% failure rate." The pipeline retrieves telemetry and clinical reports, uses a validated classifier to identify true device failures, and calculates the failure rate. One day, the pipeline executes and returns a result showing a failure rate of 0.1\%. Because the pipeline has a strong \(\mathsf{Cert}_\delta(\pi_{safe})\), its output is highly reliable. The resulting low score, \(s^{\mathrm{LB}}_{\pi_{safe}}(\varphi) \ll \theta_{AK}\), means the proposition of safety is false. The corporation now has actual knowledge of a potential safety defect. This knowledge formally triggers a duty to investigate, report, and potentially recall the product. The failure to act on this algorithmically-generated knowledge would no longer be mere negligence but would rise to the level of recklessness.

\subsection{Constructive knowledge}
In contrast to actual knowledge, which requires an epistemic act, constructive knowledge is established by the epistemic potential of the corporate mind. It is a legal artifice designed to prevent entities from benefiting from their own negligence or poorly designed information systems. Doctrines of constructive knowledge impute knowledge to a party that had a reasonable opportunity and the technical capacity to know a fact but failed to do so. The legal question is whether the corporation possessed the latent capability to know a fact, making its ignorance unreasonable. This shifts enquiry from what the corporation did know to what it ought to have known, a question our method above can measurably assess. This concept maps directly to our organisational knowledge metric, \(S_S(\varphi)\) (Eq. \ref{eq:org_score}), which represents the highest achievable score across all pipelines available within the corporation's extended cognitive architecture. Constructive knowledge can be inferred when this optimal achievable score for a proposition meets a context-appropriate, normative threshold of reasonableness (\(\theta_{CK}\)):
\[ S_S(\varphi) = \sup_{\pi\in\Pi}\; s_\pi(\varphi) \ge \theta_{CK}. \]
This inequality asserts that at least one pipeline existed within the firm's information system that could have been executed to produce timely and reliable knowledge. The components of this hypothetical optimal pipeline - its relative expected low cost \(\mathbb{E}[T_{\pi^*}]\) and low total error \(\varepsilon^{\mathrm{tot}}_{\pi^*}\) - render the act of querying (or prompting) to obtain the salient knowledge feasible. A failure to act means the firm neglected a state where the knowledge predicate would have been satisfied. It implies that a state where \(\mathsf{K}_S(\varphi;\theta_{CK}) = 1\) was readily achievable. In our approach, culpability of constructive knowledge arises directly from this inaction in the face of demonstrable epistemic capacity.

Furthermore, systemic failures that give rise to constructive knowledge can be identified through the epistemic capacity index, \(\mathcal{K}_{S,t}(\theta)\) (Eq. \ref{eq:capacity_index}). This firm-wide metric provides a prospective way for courts and regulators to assess the overall adequacy of a corporation's information systems. If a firm consistently maintains a low \(\mathcal{K}_{S,t}\) for a class of legally salient facts (e.g., safety, anti-discrimination, or anti-fraud risks), it demonstrates a systemic deficiency in its cognitive architecture. Doing so would then be not just an isolated failure to know one fact, but a broad incapacity to stay informed about its most critical obligations. Such a finding could strengthen the case for constructive knowledge across that entire class of propositions and could serve as \textit{prima facie} evidence of a failure of board oversight under the standards set in cases such as \textit{In re Caremark} and subsequent decisions, which require directors to ensure the implementation of reasonable information and reporting systems \cite{caremark1996, stone2006ritter}. A low \(\mathcal{K}_{S,t}\) contributes quantitative evidence that such systems may, in fact, be unreasonable. We set out some examples below.

\paragraph{Financial Fraud Detection}
A mid-sized bank uses a legacy, rule-based system for detecting money laundering. Its competitors have widely adopted more sophisticated graph-based neural network pipelines that are known to be far more effective. The bank has the data and the technical talent to implement a similar, high-scoring pipeline (\(S_S(\varphi_{fraud}) \ge \theta_{CK}\)), but has not done so due to cost-cutting. When a large-scale money laundering scheme goes undetected, a regulator could argue for constructive knowledge. The argument would be that the bank's epistemic capacity index, \(\mathcal{K}_{S,t}\), for the legally salient proposition of "detecting complex fraudulent transactions" was unreasonably low compared to the industry standard. The high score of the available but un-implemented pipeline demonstrates that knowledge was within reach.

\paragraph{Algorithmic Bias in Hiring} A technology company uses an AI tool to screen resumes. Academic research and open-source audit tools have demonstrated that this type of model often exhibits gender bias \cite{noble2018algorithms}. A straightforward pipeline (\(\pi_{audit}\)) exists to test the proposition (\(\varphi_{bias}\)): "Our hiring tool does not produce a disparate impact on protected groups." This pipeline, using standard statistical tests like the 80\% rule, would have a low cost and high reliability (\(S_S(\varphi_{bias})\) would be high). The company fails to implement this routine audit. When a class-action lawsuit is filed alleging discrimination, the plaintiffs can argue for constructive knowledge of the bias \cite{barocas2016big}. The company cannot claim ignorance when a simple, high-scoring pipeline to discover the truth was readily available. Its failure to run the pipeline demonstrates a failure of the due care required by employment law, making its ignorance of the system's discriminatory impact a legally culpable state.

\subsection{Wilful blindness}
The doctrine of wilful blindness applies when a firm, aware of a high probability of a fact, takes deliberate and affirmative steps to avoid confirming it. This is distinguished from a passive failure to inquire (negligence) or a failure to implement reasonable systems (constructive knowledge) by an intentional choice to remain ignorant. Our framework provides a way to attribute quantitative meaning to it by identifying the specific technical actions that constitute such deliberate avoidance. In our approach, wilful blindness is evidenced by the existence of a low-cost pipeline \(\pi_c\) with a very high potential score \(s_{\pi_c}(\varphi)\) approaching 1. It arises from two conditions:
\begin{enumerate}
    \item \textit{High efficiency}: The pipeline has a trivial cost, \(\mathbb{E}[T_{\pi_c}] \ll \tau^\star\), meaning the efficiency term \(f_\tau(\cdot)\) is nearly 1. The knowledge is, for all practical purposes, instantaneously accessible; and/or
    \item \textit{High Reliability}: The pipeline has minimal or vanishingly small total error, \(\varepsilon^{\mathrm{tot}}_{\pi_c} \approx 0\), meaning its output would be nearly certain.
\end{enumerate}
Under these conditions, the organisational knowledge metric \(S_S(\varphi)\) is therefore very high (\(\approx 1\)), meaning a state where the knowledge predicate \(\mathsf{K}_S(\varphi;\theta_C) = 1\) is trivially accessible. The existence of such a pipeline aligns with legal duties to execute it, as any reasonable entity, once aware of a high-probability risk, would take the simple, costless step to confirm it. Determining a case of wilful blindness would then focus on evidence that the corporation deliberately chose not to execute this pipeline, or actively structured its systems to avoid triggering it. Doing so would represent a conscious effort to prevent the knowledge predicate from being formally met. This is not a failure of the extended cognitive system, but a deliberate undermining of it. Such actions might include:
\begin{itemize}
    \item \textit{Technical Sabotage}: Disabling a specific search index that is known to contain incriminating data, filtering out alerts from a well-calibrated compliance model, or placing highly relevant documents in a data silo that is explicitly excluded from company-wide search procedures.
    \item \textit{Deliberate Misconfiguration}: Intentionally avoiding the simple validation and calibration steps (discussed above) that would predictably reveal a model's flaws. For instance, refusing to run a simple cross-validation test (Eq. \ref{eq:cv}) or ignoring a high \(\widehat{\mathrm{ECE}}\) score (Eq. \ref{eq:ece}) from a calibration check constitutes a deliberate choice to rely on an unreliable procedure.
\end{itemize}
This framing of wilful blindness operationalises the associated legal standards, such as those set out in \textit{United States v. Jewell}, where the court established that a defendant's suspicion and deliberate avoidance of knowledge could substitute for actual knowledge \cite{jewell1976}. This approach was affirmed in \textit{Global-Tech Appliances, Inc. v. SEB S.A.}, where the Supreme Court held that a wilfully blind defendant must take deliberate actions to avoid learning of a fact and that the fact must be highly probable \cite{globaltech2011}. Our criteria for wilful blindness aims to  quantifiable embodiment of highly probable and easily verifiable facts. We set out some examples below.

\paragraph{Sanctions Evasion} Consider a financial services firm notified by a regulator that certain transaction patterns are highly indicative of attempts to evade sanctions. The firm has a sophisticated monitoring system, and a compliance officer could easily create a simple query (\(\pi_{cheap}\)) to search for these specific patterns in their transaction logs. The query would run in seconds and have near-perfect accuracy (\(s_{\pi_{cheap}}(\varphi) \approx 1\)). Instead of running the query, a senior manager instructs the compliance team to avoid running any new proactive screens until our next scheduled audit cycle. This instruction is a deliberate action to avoid learning, and when the firm is later found to have processed sanctioned transactions, the existence of a test which the firm chose to not run would arguably constitute evidence of wilful blindness.

\paragraph{Pharmaceutical Safety} During a clinical trial for a new drug, a data scientist notices an anomaly in patient data that suggests a potential, serious side effect. To confirm this, a simple statistical test (\(\pi_{test}\)) could be run on the full dataset. The test is computationally trivial and highly reliable. The data scientist raises this to her manager, who, fearing the results could derail the drug's FDA approval, instructs her to stick to the pre-approved analysis plan and not go chasing statistical ghosts. This is a direct order to avoid executing a feasible test. The organisational knowledge metric \(S_S(\varphi_{side\_effect})\) was nearly 1, but the company deliberately chose not to complete the final epistemic step. If the drug later causes harm due to this side effect, the company cannot claim it did not know. Rather, the metrics ought to indicate that it chose not to know.

\subsection{Recklessness and negligence}
While wilful blindness involves the deliberate avoidance of a high-scoring, truth-revealing pipeline, recklessness and negligence concern the deployment of epistemically weak systems. Such epistemic states are not about avoiding knowledge, but about an unjustifiable disregard for the risks of ignorance or error. Our framework can be applied to distinguish between the conscious disregard of a known risk (recklessness) and a systemic failure to exercise due care (negligence) as follows.

\paragraph{Recklessness}
Recklessness involves the conscious disregard of a substantial and unjustifiable risk in a way that constitutes a gross deviation from the standard of conduct \cite{mpc1985}. In our framework, this translates to the act of executing a pipeline \(\pi\) despite possessing clear evidence of its unreliability. The relevant artefact for assessing recklessness is the proposed validation certificate for a pipeline, \(\mathsf{Cert}_\delta(\pi)\). A corporation acts recklessly when it proceeds even though the certificate shows the pipeline's certification score is dangerously low, \(s^{\mathrm{LB}}_\pi(\varphi) \ll \theta_R\), for salient information \(\varphi\). This low score could result from a high measured error rate (\(U_\delta(\varepsilon^{\mathrm{tot}}_\pi)\) is large) or an unacceptably long runtime. Another example would include deployment of a high-stakes AI system without any validation certification procedure at all. Such failures to perform even minimal validation, such as checking for poor calibration or assessing performance on out-of-distribution data, would arguably provide evidence towards gross deviation from the technical and ethical norms of the field \cite{asa2018ethical, acm2018code}, in this case the use of sophisticated AI and information processing systems for decisional purposes. For example, an autonomous vehicle company that deploys a perception model in snowy conditions, knowing its \(\mathsf{Cert}_\delta(\pi)\) was generated exclusively on sunny-weather data and shows a high error rate (\(U_\delta(\varepsilon^{\mathrm{ver}}_\pi)\)) in simulated winter conditions, is acting recklessly. The company is consciously disregarding the known limitations and unjustifiable risks of its epistemic procedure.

\paragraph{Negligence}
Negligence is a lesser standard than recklessness \cite{salmond1907torts}. It is characterised not by a conscious disregard of a specific risk, but by a systemic failure to maintain a reasonable standard of care. This would be a failure of the corporate cognitive architecture itself. Negligent systemic failure could be measured by the firm-wide epistemic capacity index, \(\mathcal{K}_{S,t}(\theta)\). Under this proposal, a corporation is negligent when it fails to invest in the systems, processes, and validation necessary to maintain a reasonably high \(\mathcal{K}_{S,t}\) for legally salient risks, in contravention of governance expectations e.g. like the NIST AI RMF \cite{nist_rmf_2023}. A low \(\mathcal{K}_{S,t}\) may provide a quantitative evidentiary indicator of a systemic deficiency, showing the corporation lacks the basic capacity to stay informed. For example, consider a medical diagnosis AI trained primarily on data from one demographic group. The company fails to build and implement the standard pipelines (\(\pi_{audit}\)) needed to test for bias and performance across other groups. As a result, its \(\mathcal{K}_{S,t}\) metric for the proposition such as "our model is fair and effective for all populations" would be near zero. This is not a conscious decision to deploy a known underperforming model, but a negligent failure to build the necessary systems to discover its flaws. This systemic epistemic weakness constitutes a failure of due care, making the resulting harms legally cognisable.





\section{Conclusion}\label{sec:conclusion}
In this work, we have advanced a formal thesis that corporate knowledge, in an era of AI-augmented cognition, can be defined and measured as a function of epistemic efficiency and reliable validity. By treating the corporation's information systems as part of an extended cognitive architecture, assessments of corporate epistemic states can potentially move beyond qualitative inference to quantitative measures. Our core metrics - the organisational knowledge metric \(S_S(\varphi)\), the thresholded knowledge predicate \(\mathsf{K}_S(\varphi;\theta_C)\), and the firm-wide epistemic capacity index \(\mathcal{K}_{S,t}\) provide one prospective means of rendering the corporate mind in an era dominated by AI technologies tractable. Our proposed constructs provide the basis for developing techniques for stakeholders, including corporations, courts, regulators, and boards to assess corporate \textit{mens rea} \cite{fisse1991attribution}. Importantly, the techniques complement and fit within existing legal doctrines governing epistemic states of corporations. They also align in principle with the commonplace judicial and regulatory information standards \cite{jewell1976, caremark1996, daubert1993, fre702_2023}.
\\
\\
Our framework aims to assist making claims of corporate knowledge falsifiable and auditable. It grounds abstract legal duties in concrete, measurable artefacts, such as the validation certificates, which are in turn derived from standard statistical procedures like cross-validation and calibration \cite{stone1974cv, guo2017calibration}. Ultimately, how a firm chooses to configure and validate its information pipelines is a direct reflection of its choice of how to know. Nevertheless, by deliberately measuring the capabilities of these systems, we can assist in creating more rational and predictable incentives for corporations to build systems that are not only more powerful but also more accountable.
\\
\\
Future research following on from our work has applications in three particular areas that dovetail with emergent research on AI infrastructure and multi-agent systems. The first is the empirical testing and simulation of our method, including testing the grounding of the knowledge thresholds (\(\theta_C\)) by analysing regulatory guidance and enforcement practices from bodies like the SEC and under regimes like the EU AI Act \cite{sr11_7_2011, eu_ai_act_2024}. The second is the development of verifiable governance systems, potentially using reliable credentialling technology (e.g. using cryptographic attestations and secure logging) to create trustworthy audit trails for discovery and regulatory review \cite{frcp37e_2024,kroll2017accountable}. Finally, research may explore how our framework could be extended to incorporate more complex forms of cognition, such as the multi-step, chain-of-thought reasoning now possible with advanced language models, allowing for a richer assessment of a corporation's epistemic duties.

\begin{acks}
No funding or support was received in preparation of this manuscript. The author declares no conflicts of interest. LLMs were used in the editing of this work.
\end{acks}


\bibliographystyle{plain}
\bibliography{refs2}

\appendix

\section{Simulation of Corporate Epistemic Capacity}

\subsection{Experimental Design and Objectives}
In this section, we detail the design and results of toy computational simulations created to provide demonstrations of the theoretical framework developed in the main body of the paper. The primary objective of the experiment is to operationalise our core framework that corporate knowledge in the age of AI is a measurable function of epistemic efficiency (the speed of information access) and validated reliability (the reliability of the process). We used LLMs to assist with coding the experiments. The anonymised repository of code is available via \cite{operationaliseextendedcog}.
\\
\\
To test the proposed approach from the main paper, we simulated and compared two archetypal firms:
\begin{enumerate}
    \item \textit{LegacyCorp}: Represents a company with outdated information systems. Its epistemic procedures are modeled as a slow, keyword-based search of its internal data.
    \item \textit{ModernCorp}: Represents a company that has invested in modern AI infrastructure. Its epistemic procedures are modeled as a highly efficient semantic search coupled with a powerful Large Language Model (LLM) for analysis and verification.
\end{enumerate}
By subjecting both firms' information infrastructure to a series of legally salient queries, we show how their underlying technological capabilities produce quantifiable differences in their pipeline scores (\(s_\pi(\varphi)\)), organisational knowledge metrics (\(S_S(\varphi)\)), and ultimately, their firm-wide epistemic capacity indices (\(\mathcal{K}_{S,t}\)). The simulation is designed to create clear, differentiated outcomes that directly map onto the legal doctrines of actual knowledge, constructive knowledge, wilful blindness, and recklessness.

\subsection{Methods}
The simulation was conducted in a Python environment using the Google Colab platform using API calls to OpenAI's GPT4 model. The methodology comprises four key stages:
\begin{enumerate}
    \item \textit{Corpus Creation}: A synthetic corporate data store was created, comprising 62 documents. This corpus was designed to be a realistic proxy for a company's internal knowledge base (e.g., emails, memos, reports). Specific incriminating documents were written using subtle corporate euphemisms (e.g., "market harmony" instead of "price fixing"), making them discoverable by conceptual semantic search but likely to be missed by a literal keyword search. This corpus serves as the "ground truth" for all experimental tasks.
    \item \textit{Pipeline Implementation}:
    \begin{itemize}
        \item LegacyCorp Pipeline (\(\pi_{legacy}\)): This was implemented as a simple Python function that performs a linear keyword search over the entire corpus. Its cost (\(\tau\)) was simulated to be high and proportional to the corpus size (\(O(n)\)), reflecting its inefficiency. Its primary failure mode is retrieval error (\(\varepsilon^{\mathrm{ret}}\)), which occurs if the predefined keywords for a query do not appear literally in the necessary ground-truth documents.
        \item ModernCorp Pipeline (\(\pi_{modern}\)): This pipeline first uses a pre-trained sentence-embedding model (`all-MiniLM-L6-v2`) from the Hugging Face library to convert the entire corpus into a vector space. At runtime, a query is also embedded, and retrieval is performed via highly efficient cosine similarity search. Its cost (\(\tau\)) was simulated to be low and proportional to the logarithm of the corpus size (\(O(\log n)\)). The top retrieved documents are then passed as context to a large language model (OpenAI's API) which acts as a verifier. Its primary failure mode is verification error (\(\varepsilon^{\mathrm{ver}}\)), representing a small, non-zero chance that the LLM misinterprets the context and provides an incorrect answer.
    \end{itemize}
    \item \textit{Task Execution}: A docket of four legally salient propositions (\(\Phi\)) was created, each designed to test a specific legal doctrine as discussed in Section 5. For each proposition, both the LegacyCorp and ModernCorp pipelines were executed. We measured:
    \begin{itemize}
        \item Time (\(\tau\)): The simulated wall-clock time for the pipeline to complete.
        \item Error (\(\varepsilon\)):  Determined deterministically. For LegacyCorp, \(\varepsilon^{\mathrm{ret}}\) was set to 1.0 if it failed to retrieve all ground-truth documents for a query, and 0.0 otherwise. For ModernCorp, \(\varepsilon^{\mathrm{ver}}\) was set to 1.0 if the LLM's final answer did not match the ground truth, and 0.0 otherwise.
    \end{itemize}
    \item \textit{Metric Calculation and Analysis}:
Using the measured time and error for each run, we calculated the full suite of metrics from our formal framework:
\begin{itemize}
    \item The efficiency factor, \(f_\tau(\varphi)\).
  \item The total error, \(\varepsilon^{\mathrm{tot}}_\pi\), using Eq. \ref{eq:total_error}.
\item   The pipeline score, \(s_\pi(\varphi)\), using Eq. \ref{eq:pipeline_score}.
\item   The organisational score, \(S_S(\varphi)\), using Eq. \ref{eq:org_score}, where ModernCorp's score is the supremum of its own pipeline and the available legacy pipeline.
\item  The binary knowledge predicate, \(\mathsf{K}_S(\varphi; \theta_C)\), using a normative knowledge threshold of \(\theta_C = 0.7\).
\item   The firm-wide epistemic capacity index, \(\mathcal{K}_{S,t}(\theta)\), using Eq. \ref{eq:capacity_index}.
\end{itemize}
\end{enumerate}
\subsection{Results}
The simulation produced starkly different outcomes for the two firms. Table 1 below presents the detailed performance metrics for each pipeline on each of the four legal tasks. These raw performance metrics were then used to calculate the organisational scores and the final firm-wide epistemic capacity.

\begin{table}[h!]
\centering
\caption{Detailed Simulation Performance Metrics by Legal Doctrine}
\label{tab:sim_results}
\begin{tabular}{p{1.8cm}p{2.3cm}rrrrr}
\toprule
\textbf{Company} & \textbf{Doctrine} & \textbf{Time (\(\tau\))} & \(\varepsilon^{\mathrm{ret}}\) & \(\varepsilon^{\mathrm{ver}}\) & \(\varepsilon^{\mathrm{tot}}\) & \(s_\pi\) \\
\midrule
LegacyCorp & Actual Knowledge & 5.90s & 0.00 & 0.00 & 0.00 & 0.63 \\
ModernCorp & Actual Knowledge & 2.06s & 0.00 & 0.00 & 0.00 & 0.83 \\
\midrule
LegacyCorp & Constructive & 6.05s & 1.00 & 0.00 & 1.00 & 0.00 \\
ModernCorp & Constructive & 2.09s & 0.00 & 0.00 & 0.00 & 0.83 \\
\midrule
LegacyCorp & Wilful Blindness & 5.75s & 1.00 & 0.00 & 1.00 & 0.00 \\
ModernCorp & Wilful Blindness & 2.12s & 0.00 & 0.00 & 0.00 & 0.83 \\
\midrule
LegacyCorp & Recklessness & 6.03s & 0.00 & 0.00 & 0.00 & 0.62 \\
ModernCorp & Recklessness & 2.02s & 0.00 & 0.00 & 0.00 & 0.83 \\
\bottomrule
\end{tabular}
\end{table}

\subsection{Discussion}
Although a simple toy model, the results of the simulation provide an indication of how our framework would apply:
\begin{enumerate}
    \item The most notable differences were in the constructive knowledge and recklessness scenarios. LegacyCorp's brittle keyword search failed (\(\varepsilon^{\mathrm{ret}}=1.0\)) to find documents that used non-literal, conceptual language related to hiring bias, resulting in a pipeline score of 0.0. ModernCorp's semantic search, however, easily found these documents and achieved a high score of 0.83. This gap between LegacyCorp's performance and the easily achievable performance of ModernCorp is a quantifiable measure of what LegacyCorp should have known. Its failure was not due to a lack of data, but a lack of a reasonable epistemic procedure.
    \item Similarly, the wilful blindness scenario demonstrated how a poorly designed cheap test can be a tool for deliberate ignorance. The naive keyword search for incriminating legal terms failed, yielding a score of 0.0. A regulator could argue that relying on such a brittle search, when the true conceptual query was readily available to a modern system, constitutes a deliberate choice to avoid knowledge.
\end{enumerate}
The organisation-wide results were similarly in favour of ModernCorp. These toy experiments are obviously limited and were run with minimal experimental cost. However, they do provide an indication of the type of experimental procedures that may be run in order to begin to demarcate different classifications of corporate epistemic state.


\section{Extended Simulations}
To further validate our framework and explore its implications, we conducted a series of more extensive computational experiments. These experiments aimed to provide indications (not comprehensive evidence) of statistical robustness, scalability, and theoretical boundaries of our model of corporate knowledge. We sought to explore not just that a performance gap exists between LegacyCorp and ModernCorp, but to show how this epistemic gap is systemic, predictable, and widens under realistic corporate conditions.

\subsection{Experiment 1: Statistical Robustness and Distribution}
\begin{figure}
    \centering
    \includegraphics[width=1\linewidth]{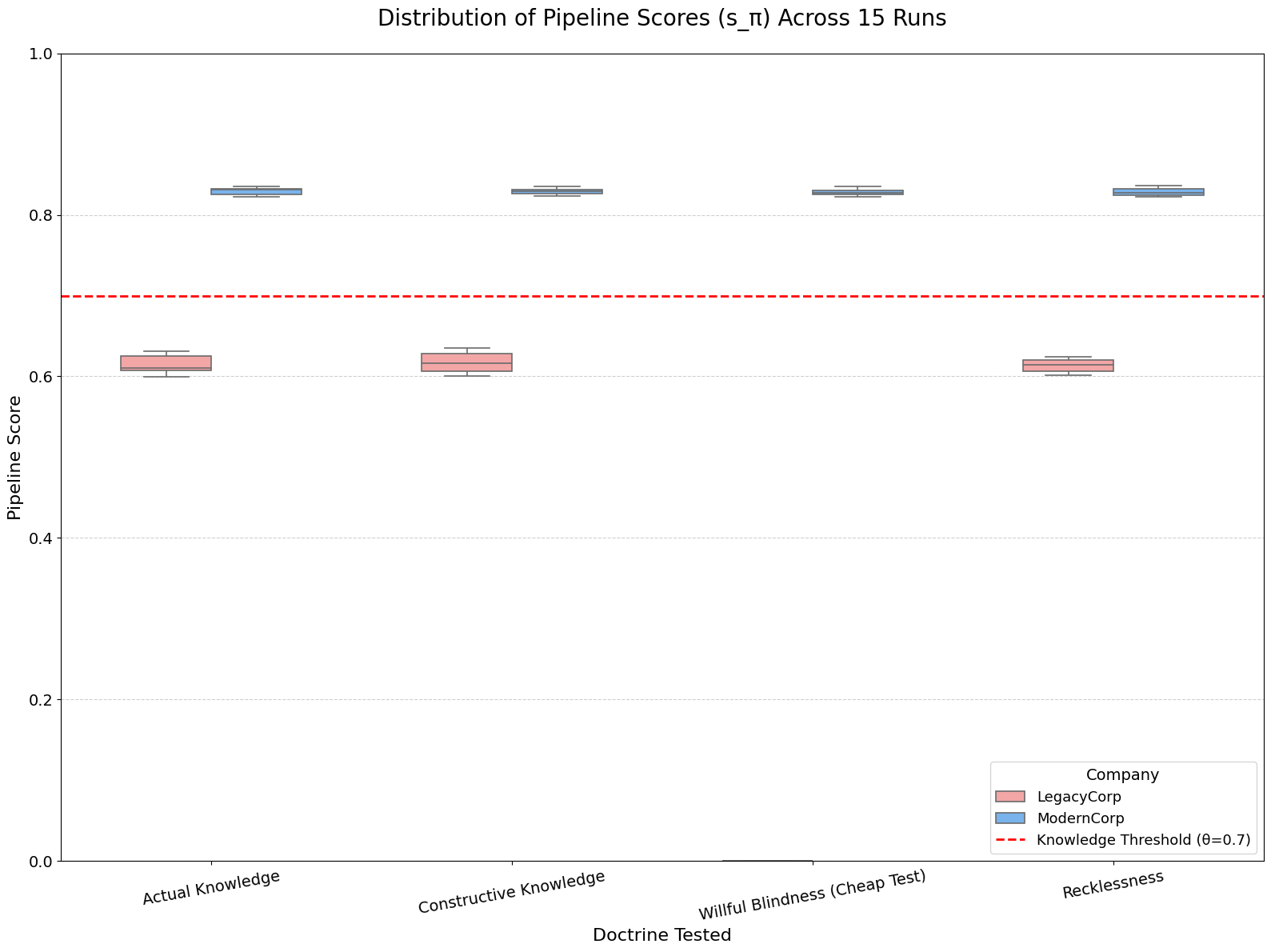}
    \caption{Distribution of pipeline scores (\(s_{\pi}\)) across 15 simulation runs for each legal doctrine. 
Each box represents the interquartile range of observed scores for a given doctrine under \textit{LegacyCorp} (pink) and \textit{ModernCorp} (blue) pipelines. 
The red dashed horizontal line marks the knowledge threshold (\(\theta_C = 0.7\)), which defines the minimum score required for a system to satisfy the formal knowledge predicate \(\mathsf{K}_S(\varphi; \theta_C)=1\). 
Across all doctrines, \textit{ModernCorp} consistently exceeds the threshold, indicating sufficient epistemic reliability and efficiency, whereas \textit{LegacyCorp} remains below the threshold, reflecting inadequate epistemic capacity. 
The stability of the boxplots demonstrates that these differences persist across repeated trials, highlighting a structural epistemic gap between the two corporate information systems.}

    \label{fig:bar}
\end{figure}
We ran a Monte Carlo simulation, running the entire experimental process from Appendix A a total of 15 times. By introducing a small amount of random variation in the simulated pipeline execution times, we aimed to observe the distribution of the pipeline scores (\(s_\pi\)) rather than a single point estimate. This was to provide a more meaningful and robust indication of the consistency and reliability of each firm's epistemic procedures. 
The results are presented in Figure \ref{fig:bar} indicating the systemic nature of each firm's epistemic capabilities:
\begin{itemize}
    \item ModernCorp's performance is characterised by tight, stable box plots consistently located well above the knowledge threshold (\(\theta_C=0.7\)). This indicates that its epistemic procedures are not only effective but also highly reliable and predictable. The narrow inter-quartile range is the quantitative signature of a robust, well-engineered cognitive system.
\item LegacyCorp's performance is, by contrast, highly volatile and task-dependent. For the actual knowledge and recklessness tasks where its brittle keyword search succeeded, the distribution of its scores is clustered well below the knowledge threshold, with the entire interquartile range failing to meet the legal standard. For the constructive knowledge and wilful blindness tasks, its performance collapses to a score of 0.0 in every single run, albeit this is an artifice of how the measures have been coded.
\end{itemize}
These types of experiments could in principle (though we did not here do this) be developed in a way that could reflect whether ModernCorp's procedures are reliable under relevant legal standards, while LegacyCorp's are demonstrably unreliable and deficient. The failure of LegacyCorp - either due to inefficiency or failing due to retrieval error - paints a picture of a system that is fundamentally unfit for the epistemic duties of a modern corporation.

\section{Experiment 2: Scalability and Epistemic Gaps}
A core proposal of this paper is that the epistemic advantage of modern AI systems grows with the scale of corporate data. To test this, we conducted a scalability analysis. We measured the time-to-knowledge (\(\tau\)) for both LegacyCorp's linear search and ModernCorp's logarithmic-time semantic search as we artificially increased the size of the corporate corpus from 60 documents to 1,000. We used a constructive knowledge query, as it represents a realistic discovery task.
\begin{figure}
    \centering
    \includegraphics[width=\linewidth]{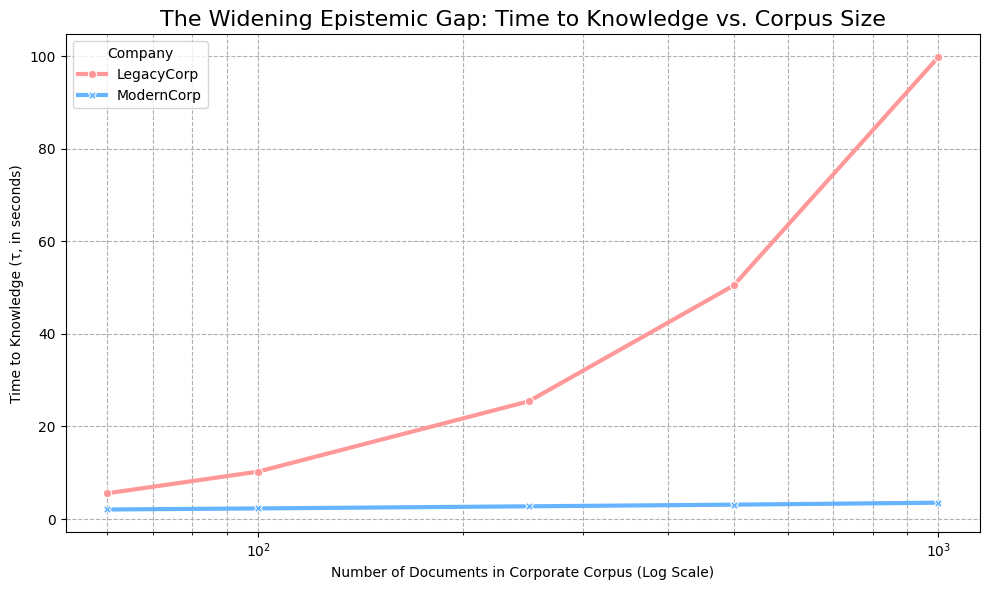}
    \caption{Epistemic scalability: comparison of time-to-knowledge (\(\tau\)) growth rates for \textit{LegacyCorp} and \textit{ModernCorp} as corporate corpus size increases. 
\textit{LegacyCorp} exhibits linear scaling (\(O(n)\)), with time-to-knowledge rising steeply from approximately six seconds for 60 documents to over 100 seconds for 1,000. 
\textit{ModernCorp}, by contrast, achieves logarithmic-time scaling (\(O(\log n)\)), maintaining nearly constant retrieval time even as the dataset expands. 
This widening gap illustrates how algorithmic efficiency compounds epistemic advantage: as data volumes grow, efficient architectures expand the feasible set of knowable facts. 
Legally, this implies that failure to adopt scalable search architectures may render a firm's ignorance increasingly unreasonable, reflecting an emergent “epistemic–culpability gap” aligned with its technological lag.}
    \label{fig:epistemicgap}
\end{figure}
The results are set out in Figure \ref{fig:epistemicgap}:
\begin{itemize}
    \item LegacyCorp's time-to-knowledge (\(\tau\)) exhibits clear growth. It starts at ~6 seconds for 60 documents and ends at over 100 seconds for 1,000. Extrapolating to a realistic corporate corpus of millions of documents, its search time would become weeks or months, rendering its epistemic procedures functionally infeasible.
\item ModernCorp's time-to-knowledge remains nearly flat. The time taken to search 1,000 documents is almost identical to the time taken to search 60.
\end{itemize}
The results provide a toy model towards how claims of ignorance on the basis that finding a specific piece of information would be like finding a needle in a haystack might be quantifiably adjudicated. At least for the scale and simple parameters of the experiment, it suggests that for a firm with modern AI, the size of accessible corpora of information is of one or more orders of magnitude, reflective of a considerable epistemic gap. The widening epistemic gap shown in the plot is therefore prospectively correlated with a wider culpability gap. Moreover, potentially as data grows, the failure to adopt efficient search technology becomes an increasingly unreasonable, and therefore more legally culpable, choice. In each case, it reiterates that a board's duty to maintain reasonable information systems must be evaluated against the technological frontier, and this chart shows that the frontier has moved dramatically.

\subsection{The Price of Unreliability}
According to our framework, efficiency alone is insufficient to constitute knowledge. The process must also be reliable. To illustrate this, we conducted a toy sensitivity analysis on ModernCorp's verifier. We held its highly efficient retrieval time constant and parametrically increased its verification error rate (\(\varepsilon_{ver}\)) from 0\% to 50\%, plotting the resulting pipeline score (\(s_\pi\)).
\begin{figure}
    \centering
    \includegraphics[width=\linewidth]{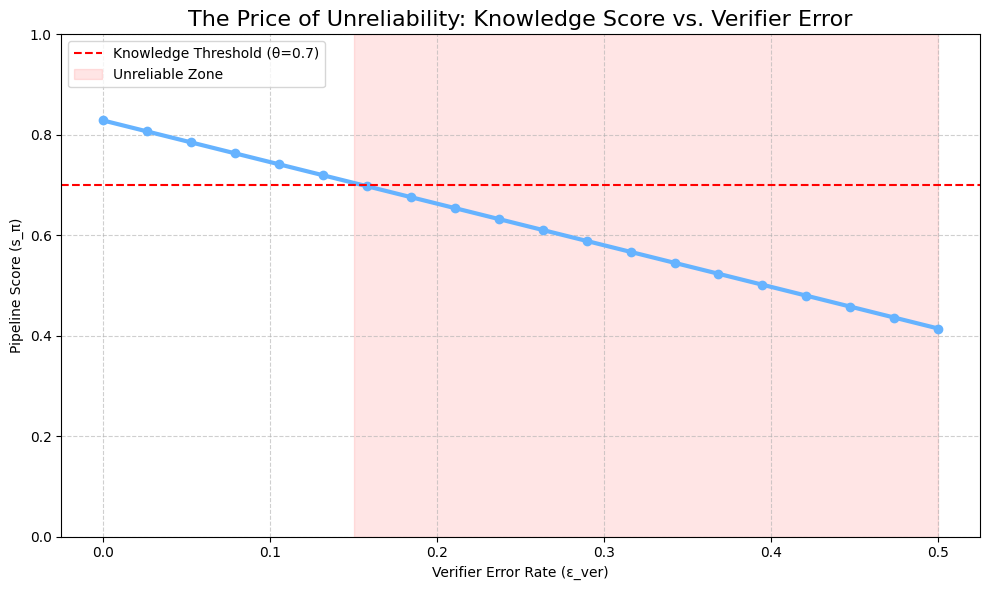}
    \caption{The price of unreliability: degradation of the pipeline score (\(s_{\pi}\)) as the verifier error rate (\(\varepsilon_{\mathrm{ver}}\)) increases. 
The blue line traces how the expected knowledge score declines with rising verification error, holding retrieval efficiency constant. 
The red dashed line marks the knowledge threshold (\(\theta_C = 0.7\)) required for epistemic sufficiency, while the shaded region indicates the ``unreliable zone,'' where verifier error rates above 15\% drive the system below the legal knowledge threshold. 
The plot demonstrates that even highly efficient systems lose epistemic adequacy once reliability falls below this critical point, emphasizing that validation and calibration of AI verifiers are indispensable for legally cognizable knowledge.}

    \label{fig:price}
\end{figure}
Results are visualised in Figure \ref{fig:price}. As the verifier's error rate increases, the pipeline score for ModernCorp steadily degrades, even though its retrieval is instantaneous. The score crosses the legal knowledge threshold (\(\theta_C=0.7\)) at an error rate of approximately 15\%. This result indicates the importance of the validation component of our framework, as detailed in Section 4. A corporation cannot simply deploy a fast, efficient AI and claim it has knowledge. It has a continuing duty to validate that the AI's outputs are reliable. A failure to perform the necessary calibration and testing to keep \(\varepsilon_{ver}\) low can render an otherwise powerful system epistemically worthless and legally indefensible. This is directly consistent with jurisprudential standards (e.g. \textit{Daubert}) for a "known error rate," showing that a high error rate, even for a technologically advanced system, can be disqualifying.

\end{document}